\documentclass[11pt]{article}

\usepackage[preprint]{acl}

\usepackage{times}
\usepackage{latexsym}

\usepackage[T1]{fontenc}
\usepackage[utf8]{inputenc}

\usepackage{microtype}

\usepackage{inconsolata}

\usepackage{graphicx}

\usepackage{blindtext}
\usepackage{booktabs}
\usepackage{multirow}
\usepackage{xcolor}
\usepackage{colortbl}
\usepackage{subcaption}
\usepackage{multirow}
\usepackage{verbatim}
\usepackage{amsmath}
\usepackage{todonotes}
\usepackage{float}

\title{
Evaluation Pitfalls and Sparsity Limitations in LLM-based\\Confidence Estimates for Classification
}

\author{Elena Merdjanovska\thanks{Work done during an internship at Amazon.}\\
  Humboldt-Universität zu Berlin \\
  Science of Intelligence \\
  \texttt{\normalsize elena.merdjanovska@hu-berlin.de}\\\And
  Omar Zaidan \\
  Amazon \\
  \texttt{\normalsize ozaidan@amazon.de}\\\And
  Andreas Rücklé \\
  Amazon \\
  \texttt{\normalsize arueckle@amazon.de}\\}

\begin{document}
\maketitle
\begin{abstract}
Confidence estimation is essential when LLMs are used for classification, indicating when predictions can be trusted. However, common approaches such as verbalization produce extremely sparse outputs. For instance, Qwen3-32B verbalizes only eight unique confidence values on SST-2, with over half being exactly 95\%---a pattern we observe consistently across four datasets and two LLMs. Besides limiting practical utility, we show that this sparsity critically affects evaluation: the choice of interpolation in area under the accuracy-rejection curve (AUARC) dramatically alters rankings, with consistency sampling dropping from best to worst under stepwise versus linear interpolation. We advocate for standardizing stepwise interpolation for a fairer comparison. Under such a fair evaluation, we find that weighting verbalized digits by token probabilities---a method we term \emph{verbalization logprobs}---addresses sparsity and achieves the best AUARC (+2.3 points over vanilla verbalization) without incurring additional inference cost.
\end{abstract}

\section{Introduction}

LLMs are increasingly used for classification tasks, e.g., automatic evaluation~\citep{10.5555/3666122.3668142,lee-etal-2025-checkeval}, content moderation~\citep{yin2025bingoguard,nghiem2025smarter}, and more~\citep{marvin2025scaling,10.1145/3696630.3728552}. %For many applications, we require confidence estimates---e.g., to filter out low-confidence predictions~\citep{chen-etal-2023-adaptation,pmlr-v239-ren23a,JMLR:v11:el-yaniv10a}. 
Many applications require confidence estimates, e.g., in selective prediction, where a model can \emph{reject} to classify examples with low confidence~\citep{chen-etal-2023-adaptation,pmlr-v239-ren23a}. Such classifiers are evaluated only on the subset of examples that exceed a specific confidence threshold, with the rest left unclassified, offering control over the risk-coverage trade-off~\citep{JMLR:v11:el-yaniv10a}.

Prompting poses unique challenges for confidence estimation, as it lacks a direct mapping from outputs to class probabilities.
Novel approaches such as verbalization can prompt models to express confidence for class labels in natural language~\citep{xuan-etal-2025-seeing,liu-etal-2025-metafaith,zeng-etal-2025-thinking}, while sampling-based approaches generate multiple predictions at a higher temperature and aggregate them~\citep{phillips2025geometric,10.5555/3737916.3738199}.
These approaches differ fundamentally from traditional classifiers due to their natural language interface.
For instance, the strong preference of LLMs towards generating certain numerical tokens~\citep{coronadoblázquez2025deterministicprobabilisticpsychologyllms,shao2025benford} can bias confidence estimates, leading to LLM overconfidence~\citep{xiong2024can}. Although prior work has focused on calibration~\citep{tonolini-etal-2024-bayesian, wang2024calibrating}, we study \emph{sparsity} as a distinct and critical challenge.

%rather than investigating limitations related to sparsity. %---a limitation that, as we demonstrate, significantly impacts the practical utility of these methods.

%implications of this have not been studied in sufficient detail thus far, especially also as confidence estimation literature mostly focuses on text generation and question answering tasks~\citep{becker2024cyclesthoughtmeasuringllm, kadavath2022languagemodelsmostlyknow, xiong2024can,geng-etal-2024-survey,huang2024survey,vashurin-etal-2025-benchmarking}. Although some work has explored confidence estimation for classification~\citep{tonolini-etal-2024-bayesian, wang2024calibrating}, these studies have not investigated limitations related to sparsity. %---a limitation that, as we demonstrate, significantly impacts the practical utility of these methods.

\begin{figure*}
    \centering
    \small
    \begin{subfigure}{0.47\textwidth}
        \centering
        \includegraphics[width=\linewidth]{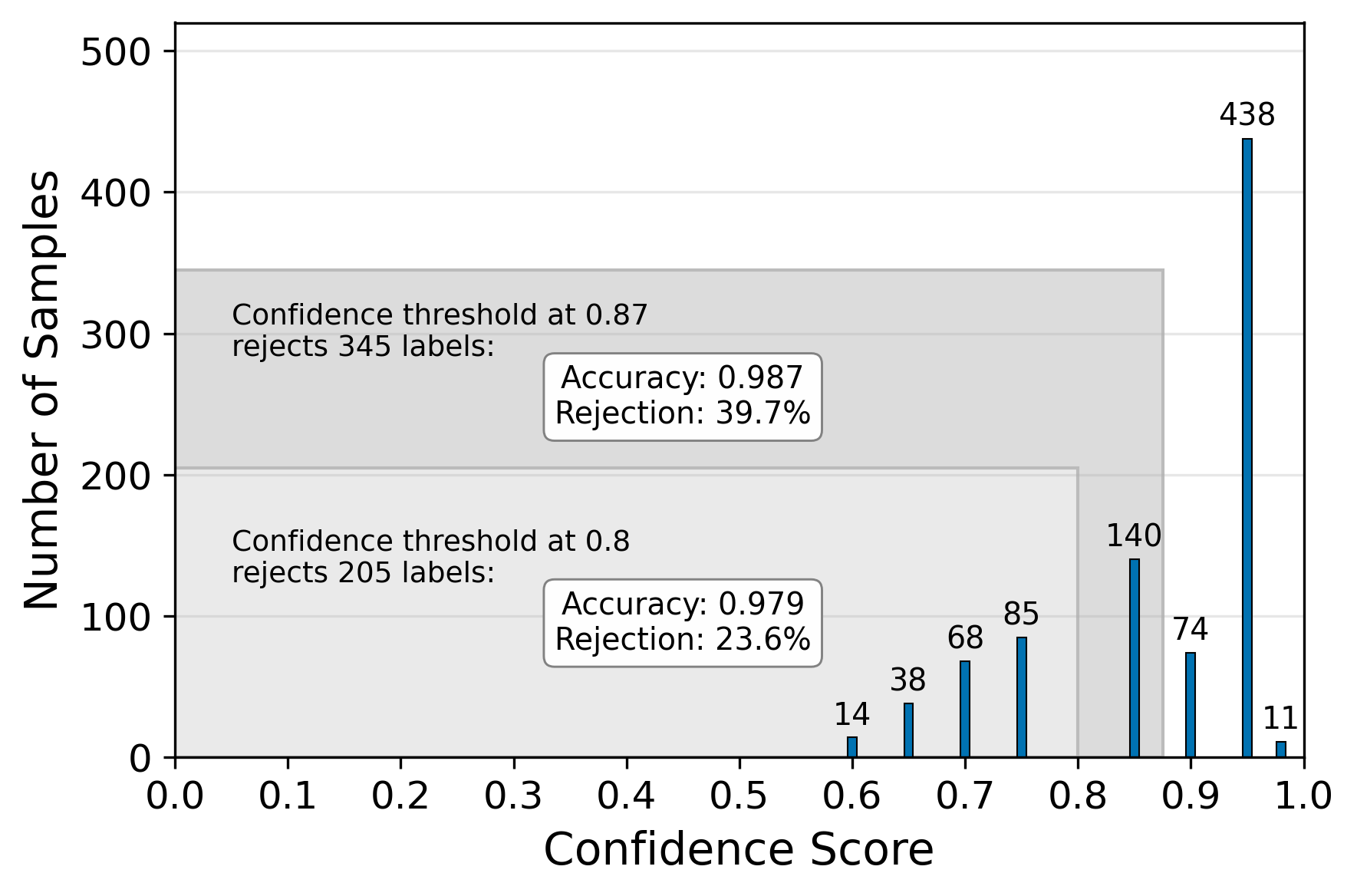}
        \caption{Histogram showing all confidence scores on SST-2}
        \label{fig:practical_sparsity_histogram}
    \end{subfigure}
    \hfill
    \begin{subfigure}{0.47\textwidth}
        \centering
        \includegraphics[width=\linewidth]{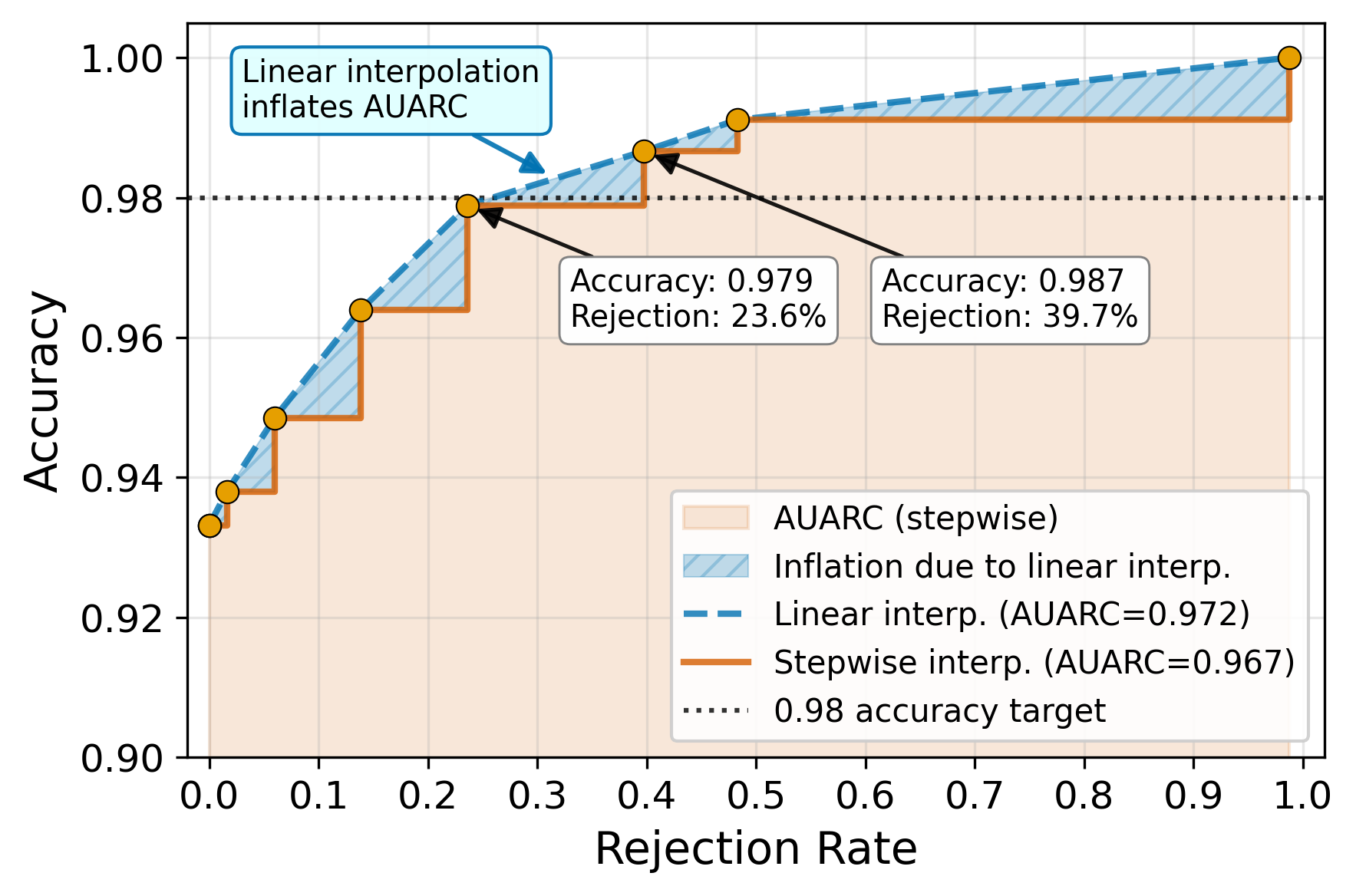}
        \caption{Accuracy-rejection curve on SST-2\footnotemark}
        \label{fig:practical_sparsity_arc}
    \end{subfigure}
    %\caption{(a) Histogram of vanilla verbalized confidences on SST-2 with Qwen3-32B. The model only predicts eight unique confidence values, namely 0.6, 0.65, 0.7, 0.75, 0.85, 0.9, 0.95, and 0.98. 
    %We show the accuracy for rejecting the lowest confidence labels for two thresholds. 
    %In (b) we show how this limits practical use: if we need to ensure our system reaches 98\% accuracy on predicted labels, we need to choose a confidence threshold that leads to a high rejection rate of 39.7\%---up from 23.6\% rejection for 97.9\% accuracy. We also illustrate how linear interpolation inflates scores: all thresholds between two points would yield the same accuracies and hence stepwise interpolation should be used.}
    \caption{\textbf{Sparse confidences limit practical utility.} 
    (a)~Histogram of vanilla verbalized confidences on SST-2 with Qwen3-32B, showing extreme sparsity with only eight unique values (0.6, 0.65, 0.7, 0.75, 0.85, 0.9, 0.95, 0.98). 
    The shaded regions highlight two rejection strategies: rejecting 205 samples (lighter gray) achieves 97.9\% accuracy, while rejecting 345 samples (darker gray) achieves 98.7\% accuracy.
    (b)~Accuracy-rejection curve demonstrating limited threshold choices due to sparsity. 
    Reaching at least 98\% accuracy requires rejecting 39.7\% of predictions vs. rejecting 23.6\% for 97.9\% accuracy (16.1pp increase). 
    Stepwise interpolation (solid orange) is correct for sparse confidences; linear interpolation (dashed blue) incorrectly assumes performance increases gradually between thresholds, creating artificial inflation (blue shaded area).}    
    \label{fig:practical_sparsity}
\end{figure*}

%In this work, we investigate the \emph{sparsity} of confidence values induced by prompted LLMs for classification tasks. 
First, we demonstrate that verbalization-based approaches produce extremely sparse confidences. For instance, Qwen3-32B~\citep{yang2025qwen3} predicts only eight unique confidence values (0.6, 0.65, 0.7, 0.75, 0.85, 0.9, 0.95, and 0.98) across SST-2~\citep{socher-etal-2013-recursive}. %, 0.95 being overwhelmingly the most common. 
This means we have \emph{little choice when selecting decision thresholds}, which limits the practical utility of such approaches. %: retaining only high-confidence predictions to achieve 97\% accuracy on that dataset would require us to reject 14.6\% of predictions, compared to just 2.2\% rejection for 96.4\% accuracy (see Figure~\ref{fig:practical_sparsity_arc}).

Second, we show that sparsity has \emph{critical implications for evaluation}. 
Using the selective classification metric AUARC~\citep{Nadeem2009}---the area under the accuracy-rejection curve---we find that a simple and often neglected aspect of metric computation, namely the interpolation method, in fact has a major impact on system rankings.
%Namely, the change of interpolation method between different confidence thresholds can change the best-performing method under linear interpolation to the worst under stepwise interpolation in our experiments.
%As literature uses these interpolations inconsistently, proper evaluation of sparse confidence methods is critical.
%
The performance of verbalization and sampling approaches decrease by 0.7--4.5 points under stepwise interpolation (vs. linear interpolation), while the continuously-distributed token logprobs approach remains unaffected.
Most strikingly, the best approach using linear interpolation in our experiments becomes the worst using stepwise interpolation.
%---consistency sampling drops 12.5 points.
%As literature uses interpolations inconsistently, our experiments illustrate the importance of considering effects from sparsity.
Given the inconsistent usage of interpolation methods in the literature, we advocate for standardizing stepwise interpolation for comparable and fair evaluations.

%On the example of AUARC~\citep{Nadeem2009}---the area under the accuracy-rejection curve---we also reveal that the sparsity of LLM-based confidence estimates has critical implications for evaluation methodology. 
%While the method of interpolation between different confidence thresholds has limited to no impact when dealing with many possible thresholds, 
%We find that the method of interpolation between performance values at different confidence thresholds can substantially alter performance rankings. Our experiments across four datasets demonstrate that consistency sampling shifts from the best-performing method under linear interpolation to the worst under stepwise interpolation. As literature does not use the same interpolations consistently, our experiments illustrate the importance of considering effects from sparsity. %when using confidences from prompted LLMs.

%computed using linear interpolation between all possible threhsolds~\citep{fix,me}, which has little to no impact on traditional methods. However, for sparse confidence distributions the more appropriate stepwise interpolation produces substantially different results. Specifically, verbalization approaches show performance drops of 0.7--4.5 points on AUARC when evaluated properly, with consistency sampling experiencing a dramatic 12.5-point decrease. This correction fundamentally alters performance rankings---consistency sampling shifts from the best-performing method under linear interpolation to the worst under stepwise interpolation, highlighting the importance of appropriate evaluation protocols.

Third, we address sparsity directly: we propose \emph{verbalization logprobs}, which weighs verbalized confidence digits by their token probabilities, transforming sparse outputs into continuous distributions.
Across four datasets, verbalization logprobs achieves +2.3 AUARC points over vanilla verbalization at no additional computational cost.

The remainder of this paper reviews existing approaches~(§2), analyzes the sparsity limitation~(§3), and proposes verbalization logprobs~(§4). %, and concludes~(§5).

\begin{table}
    \centering
    \small
    \begin{tabular}{lp{3.90cm}}
        \toprule
        \multicolumn{2}{c}{\textbf{Verbalization-based}} \\
        \midrule
        Vanilla verbalization & Direct self-reported confidence \\
        \arrayrulecolor{gray!30}\cmidrule{1-2}\arrayrulecolor{black}
        Top-k verbalization & k guesses with probabilities \\
        \midrule
        \multicolumn{2}{c}{\textbf{Sampling-based}} \\
        \arrayrulecolor{gray!30}\midrule\arrayrulecolor{black}
        Consistency sampling & Frequency of class predictions \\
        \arrayrulecolor{gray!30}\cmidrule{1-2}\arrayrulecolor{black}
        Verbalization sampling & Mean of verbalized confidences \\
        \midrule
        \multicolumn{2}{c}{\textbf{Logit-based}} \\
        \arrayrulecolor{gray!30}\midrule\arrayrulecolor{black}
        Token logprobs & Probability of class label token \\
        \bottomrule
    \end{tabular}
    \caption{Overview of the five confidence estimation methods studied in this work.}
    \label{tab:methods-overview}
\end{table}

\footnotetext{All accuracy-rejection curves have an accuracy of 1.0 at rejection rate 1.0 by convention~\citep{Nadeem2009}.}

\section{Background: Confidence Estimation for LLM Classifiers}

We distinguish three types of confidence estimation approaches for classification tasks, (1)~verbalization-based, (2)~sampling-based, and (3)~logit-based.\footnote{Note that further sub-categorization exists for generative tasks. See recent surveys by~\citet{geng-etal-2024-survey} and~\citet{liu2025uncertainty}.}
Additionally, several white-box approaches~\citep{vashurin-etal-2025-benchmarking,vazhentsev-etal-2025-unconditional} leverage internal model states for confidence estimation, which we do not study in this work. While this presents an interesting line of work, such approaches require access beyond what most widely-used LLM APIs provide.
Table~\ref{tab:methods-overview} shows the approaches we study in \S\ref{sec:sparsity-limitation}---due to their wide adoption and good reported performances---with their characteristics highlighted below.

\paragraph{Verbalization-based methods.} 
The simplest approach is prompting the model to output self-reported confidence, termed \textit{vanilla verbalization}~\citep{xiong2024can}.
%The most straightforward approach to LLM-based confidence estimation is to prompt the model to output self-reported confidence scores, which we refer to as \emph{vanilla verbalization}~\citep{xiong2024can}.
One may use different verbalized number ranges, but typically we ask the LLM to produce a confidence from 0 to 100. We can also use prompting techniques such as CoT~\citep{kojima2023largelanguagemodelszeroshot}, self-probing \cite{xiong2024can} and top-k \citep{tian-etal-2023-just}. For instance, \textit{top-2 verbalization} prompts the model to produce the \emph{two} most likely predictions, each with an associated confidence value, even though only the highest-confidence prediction is used. This outperforms other verbalization approaches~\citep{xiong2024can}. Other interesting extensions are possible, e.g., task-specific rubrics~\citep{kim-etal-2024-prometheus}, but those require task-specific prompt changes. In contrast, we focus on methods that generalize across classification tasks without requiring knowledge about the task.

\paragraph{Sampling-based methods.}
Another popular approach to confidence estimation involves sampling multiple LLM responses and aggregating their outputs. \textit{Consistency sampling} generates multiple class predictions and considers their frequency as a confidence score. Typically, performance saturates quickly~\citep{duan-etal-2024-shifting, lin2024generating}, and therefore we sample four responses in our experiments. \textit{Verbalization sampling} extends this by computing the mean of sampled verbalized confidences~\citep{xiong2024can}. For question-answering tasks, more sophisticated similarity-based approaches perform well, including semantic entropy \citep{kuhn2023semantic}
and eccentricity~\citep{lin2024generating}. However, those are not applicable to simply predicting class labels.
One downside of sampling-based approaches is that they substantially increase the number of output tokens, especially when producing reasoning traces or explanations as part of the prediction.

\paragraph{Logit-based methods.}
As verbalization- and sampling-based methods make no use of model internals, they are black-box methods that only operate on the output token sequence.
In contrast, logit-based methods occupy an interesting space between black-box and white-box. Some APIs do allow access to a limited number of log-probabilities, which we can use to estimate confidence. \textit{Token logprobs} uses the probability of the class label token as a confidence estimate~\cite{xiong2024can}. Other common methods are P(True) \cite{kadavath2022languagemodelsmostlyknow} or relying on entropy of the generated sequence in open-ended generation~\cite{huang2025look}. While these methods are closest to traditional confidence estimates by directly accessing prediction probabilities, they cannot be used with APIs that don't return logprobs (e.g., OpenAI, Claude, Bedrock). This often leaves us with only verbalization or sampling approaches being applicable in practice.

\section{The Sparsity Limitation}
\label{sec:sparsity-limitation}

We now study how sparsity affects approaches from the categories outlined before. We find substantial variation and critical implications for evaluation: a methodological choice, namely interpolation, can completely reverse performance rankings.

\paragraph{Verbalized confidences are sparse.}
LLMs tend to prefer predicting certain number tokens more frequently than others~\citep{coronadoblázquez2025deterministicprobabilisticpsychologyllms,shao2025benford}. In Figure~\ref{fig:practical_sparsity_histogram}, we show this phenomenon also holds true for confidence estimation with verbalization approaches. Qwen3-32B~\citep{yang2025qwen3} with vanilla verbalization predicts just eight unique confidence values over the entire SST-2 dataset, with \emph{more than half} of the confidence values exactly equal to 95\%.
We find that this is a relatively broad phenomenon; when running different approaches on four classification datasets\footnote{See Appendix~\ref{appendix:datasets} for dataset details.}---namely SST-2, SST-5, Amazon ESCI product classification~\citep{reddy2022shopping}, Yahoo! answers topic classification~\citep{zhang2015character}---we find that 45--93\% of confidences are concentrated in the five most common confidence values (see \%t5 in Table~\ref{tab:auarc-comparison-qwen}). We observe the same behavior using Claude 3.7 Sonnet, and both models with or without reasoning enabled (see Appendices~\ref{appendix:claude-results} and~\ref{appendix:reasoning}).
We also verify that these findings hold for an alternative confidence range (0--9) in Appendix~\ref{appendix:alternative-confidence-range}.
This limits practical utility, as illustrated in Figure~\ref{fig:practical_sparsity}: achieving small accuracy gains often requires us to accept large increases in rejection rate.

%Methods based on token probabilities, on the other hand, are not affected by sparsity as two different predictions would rarely be assigned exactly the same probabilities. However, they often underperform verbalization approaches in common evaluation setups~\citep{find,some,examples}. This suggests that the standard evaluation protocol does not sufficiently incorporate effects from sparsity.

\paragraph{Evaluations need to account for sparsity.}
Many metrics rely on measuring performance at every possible confidence threshold and integrate the area under the curve. Examples are AUROC~\citep{hanley1982meaning}, PRR~\citep{vashurin-etal-2025-benchmarking}, AUPRC~\citep{ling-etal-2024-uncertainty}, and AUARC~\citep{Nadeem2009}. AUARC is closely related to use-cases where we reject low-confidence predictions~\citep[e.g.,][]{JMLR:v11:el-yaniv10a} as it computes the area under the accuracy-rejection curve.
For sparse confidence estimation approaches---where only a few thresholds produce distinct outcomes---the choice of which interpolation method to use is critical, as we show in Figure~\ref{fig:practical_sparsity_arc}.

However, we find that this choice is usually overlooked, leading to inconsistencies across evaluations due to differing interpolation methods. While the default in AUROC and AUPRC is stepwise interpolation~\citep{muschelli2020roc,chen2024commonly}, AUARC is often computed using linear (aka trapezoidal) interpolation.\footnote{We verified this by inspecting the publicly available code of various works~\citep{nguyen-etal-2025-beyond,lin-etal-2024-contextualized,lin2024generating,vashurin-etal-2025-benchmarking}. More details in Appendix \ref{sec:prior-work-implementation}.}
%and sometimes with stepwise interpolation~\citep{10.5555/3737916.3738199}. %and in some cases it remains unclear~\citep{}. 
While this has little impact on evaluations of methods with many distinct thresholds---i.e., if there is only a small distance between subsequent points---it becomes critical when dealing with sparse confidence scores. The correct way is to use stepwise interpolation, as seen in Figure~\ref{fig:practical_sparsity_arc}.

%
%Different prior work uses linear interpolation~\citep{lin2024generating,fix,me} (and their proposed methods are not affected by sparsity).
%As we have seen in Figure~\ref{fig:practical_sparsity_arc}, there are sometimes considerable step sizes between subsequent confidence thresholds---both in terms of accuracy and rejection rate---and hence we must use stepwise interpolation to fairly compare different confidence estimation methods.

\paragraph{AUARC interpolation considerably impacts evaluation ranks.}
Table~\ref{tab:auarc-comparison-qwen} shows the impact of using stepwise rather than linear interpolation when calculating AUARC for Qwen3-32B (see Appendix~\ref{appendix:claude-results} for Claude 3.7 Sonnet results). Sparse approaches show consistent and sometimes substantial performance drops, while token logprobs---which does not suffer from sparsity---remains unchanged. Specifically, verbalization approaches score 0.7--4.5 AUARC points lower with stepwise interpolation, indicating that linear interpolation typically inflates their scores. The choice of interpolation also causes dramatic rank changes: consistency sampling ranks first with linear interpolation but drops to last place with stepwise interpolation, a 12.5 point absolute decrease.

\begin{table}[]
\centering
\small
\begin{tabular}{@{}lc c c c@{}}
\toprule
 & \multicolumn{3}{c}{AUARC} & \\
 \cmidrule(lr){2-4}
& linear & & step & \%t5 \\ \midrule
\noalign{\vskip -1.5mm}
 & & {\tiny ranks} & & \\
Consistency sampling & 0.808 & {\tiny 1→5} & 0.683 & 100 \\
Vanilla verbalization & 0.758 & {\tiny 2→4} & 0.713 & 92 \\
Top-2 verbalization & 0.757 & {\tiny 3→2} & 0.731 & 93 \\
Verbalization sampling & 0.741 & {\tiny 4→1} & 0.734 & 45 \\
Token logprobs & 0.723 & {\tiny 5→3} & 0.723 & 0 \\ \bottomrule
\end{tabular}
\caption{Impact of interpolation method on AUARC scores and rankings for Qwen3-32B. Scores are averages over SST-2, SST-5, Amazon ESCI, and Yahoo! topic classification (see Table~\ref{tab:datasets_table} for dataset details).
%Linear interpolation inflates scores for sparse methods (0.7--4.5 points) while leaving token logprobs unchanged. 
%Consistency sampling drops from first to last place (rank 1→5).
\%t5 refers to the concentration in top-5 confidence values.}
\label{tab:auarc-comparison-qwen}
\end{table}

In summary, many confidence estimation approaches for prompted LLMs---particularly verbalization-based methods---produce highly sparse outputs, concentrated within a few unique confidence values. This sparsity severely limits practical utility and needs to be appropriately reflected in evaluation methodologies. Our experiments demonstrate that the choice of interpolation method for threshold-based metrics critically impacts both absolute performance scores and relative rankings. We therefore advocate for using stepwise interpolation rather than linear interpolation when reporting such metrics, as it provides fair comparisons by properly accounting for sparsity.

\section{Verbalization Logprobs}

%We conduct a brief case-study of our evaluation best-practices and show that incorporating vanilla verbalization with token logprobs can lead to an effective combination of those two orthogonal methods.
Having observed performance drops for verbalization under proper evaluation while token logprobs remains stable, we investigate whether their combination can mitigate these losses. 
Specifically, we propose incorporating token probabilities into vanilla verbalization to reduce sparsity.
We argue that vanilla verbalization discards rich information: it obtains the sampled digit but ignores the probability the model assigned to generating it. For instance, using vanilla verbalization, we might sample the sequence ``\texttt{the confidence is 95\%}'' and the standard approach is to take 95\% \emph{at face value}. However, each digit token (`\texttt{9}' and `\texttt{5}') has an underlying probability distribution over alternatives that we can leverage.

For \textit{verbalization logprobs}, we compute confidence as the expected value over possible digits at each position:
\begin{equation*}
%\label{eq:verbalization-logprobs}
%10 \cdot \sum_{x=1}^{9} x \cdot P_i(x) + \sum_{x=1}^{9} x \cdot P_{i+1}(x)
\sum_{d=0}^{9} 10d \cdot P(x_i = d) + \sum_{d=0}^{9} d \cdot P(x_{i+1} = d)
\end{equation*}
where $x_i$ and $x_{i+1}$ are the tens and units digit tokens, respectively, and $P$ is the LLM's token probability. This transforms verbalized confidences from discrete sampled values into continuous estimates.\footnote{Strictly speaking, this expectation should use conditional probabilities, since the two digit positions are not independent for autoregressive LLMs. In this paper, we treat them as independent as an approximation borne out of practical limitations: computing the exact expected value would require knowing the units digit distribution conditioned on \emph{each} possible tens digit, i.e., running a forward pass for each one, but standard API access only returns logprobs for the actually generated token. We believe this is a reasonable approximation, since the tens digit dominates the expected value anyway.}

Table~\ref{tab:main_results_qwen} shows the results using the same setup as before, with additional metrics AUROC, ECE, and cost\footnote{Relative cost based on input/output token counts; see Appendix~\ref{appendix:cost} for details.}. We find that verbalization logprobs effectively combines the best of both worlds, by avoiding the sparsity of vanilla verbalization (from 92 to 1~\%t5) and improving all metrics compared with both vanilla verbalization (+2.3~percentage points AUARC, +0.9~AUROC) and token logprobs (+1.3~AUARC, +4.7~AUROC). This suggests that verbalized digit distribution carries useful confidence signals beyond what token logprobs provides. Compared with verbalization sampling (the strongest baseline by AUARC), verbalization logprobs achieves comparable performance (0.736 vs.~0.734) at a fraction of the cost (1.7$\times$ vs.\ 6.8$\times$), as it requires only a single inference call rather than four.

\begin{table}[]
\centering
\small
\setlength{\tabcolsep}{5pt}
\begin{tabular}{@{}lccccc@{}}
\toprule
& AUROC & ECE & AUARC & \%t5 & Cost \\ \midrule
Random conf. & 0.482 & 0.223 & 0.652 & 0 & 1.0$\times$ \\ \midrule
Consist. samp. & 0.573 & 0.296 & 0.683 & 100 & 4.0$\times$ \\
Verb. sampling & 0.678 & 0.238 & 0.734 & 45 & 6.8$\times$ \\
Top-2 verb. & 0.689 & 0.202 & 0.731 & 93 & 2.3$\times$ \\
\midrule
Vanilla verb. & 0.661 & 0.235 & 0.713 & 92 & 1.7$\times$ \\
Token logpr. & 0.623 & 0.283 & 0.723 & 0 & 1.0$\times$ \\ 
\midrule
Verbalization- & & & & & \\
~~~~logprobs & 0.670 & 0.234 & 0.736 & 1 & 1.7$\times$ \\
\bottomrule
\end{tabular}
%\caption{Results for Qwen3-32B (macro-averages).}
\caption{Comparison of confidence estimation methods with Qwen3-32B (averages over our four datasets). \%t5 indicates concentration in top-5 most frequent confidence values. ECE is the expected calibration error. Cost is a multiple relative to simply predicting the class label (no confidence).}

\label{tab:main_results_qwen}
\end{table}

In addition to the above empirical evidence, showing the benefit of our method, we provide an analysis of how its confidence scores relate to the vanilla verbalization method, to lend some theoretical justification for why we would expect improved results.
To that end, we measured the correlation between the scores of the two methods. Since both methods measure the same underlying belief of the model, it is expected that they would have high correlation. The question is (a)~just how high is the correlation? (if it is ``too high'', say > 98\%, this implies the two methods are redundant), and (b)~what practical difference does it make for someone who wants to use the confidence scores for thresholding/rejection decisions?

Table~\ref{tab:correlation_analysis} shows the correlation (Spearman's $\rho$) between verbalization logprobs and vanilla verbalization across all four datasets. We find that $\rho$  averages 0.89 across the four datasets, meaning the two methods are, as expected, highly correlated but still with a clear divergence from each other.\footnote{Other divergence measures we examined were means of absolute differences, medians of absolute differences, and differences of median confidences. In all those measures we find the same pattern: the two methods are similar yet meaningfully different.}

\begin{table}[]
\centering
\small
\setlength{\tabcolsep}{5pt}
\begin{tabular}{@{}lccc@{}}
\toprule
        &        & Vanilla   & Logprobs  \\
Dataset & $\rho$ & \# unique & \# unique \\
        &        &  values   & values    \\
\midrule
SST-2   &  0.93  &      8    &     630   \\
SST-5   &  0.97  &     11    &   1,466   \\
ESCI    &  0.76  &     11    &   2,227   \\
Yahoo   &  0.90  &     12    &   1,869   \\
\bottomrule
\end{tabular}
\caption{Correlation (Spearman's $\rho$) between confidences of vanilla verbalization (``Vanilla'') and verbalization logprobs (``Logprobs'') across all four datasets (using Qwen3-32B).}
\label{tab:correlation_analysis}
\end{table}

On the point of practical impact, we note that vanilla verbalization collapses confidence predictions into at most only 12 unique values (and as few as 8 in SST-2), while verbalization logprobs uses at least 630 unique values (and as many as 2,227 in ESCI). This is an 80$\times$ to 200$\times$ increase in resolution, which is critical and has practical consequences: as shown in Figure~\ref{fig:practical_sparsity}, if we desire $\geq98\%$ accuracy, with vanilla verbalization one must go from rejecting 23.6\% to rejecting 39.7\% of predictions because there is no threshold in between. In contrast, verbalization logprobs provides the fine-grained thresholds to avoid such large jumps: one need only reject 28.1\% of examples to achieve $\geq98\%$ accuracy, thus salvaging a full 11.6\%\ of the data (39.7\% - 28.1\%).

%Verbalization logprobs achieves the best AUARC (+2.3 points over vanilla verbalization, slightly above verbalization sampling) while keeping the cost comparatively low---verbalization sampling is more expensive as it requires sampling four responses. 

\section{Conclusion}

Sparsity of confidence estimates, i.e., producing only a handful of unique confidence values, is a problem that can severely limit the practical usefulness of prompted LLMs for classification. It is hence important that evaluation metrics account for this sparsity by using stepwise interpolation, as opposed to linear (aka trapezoidal) interpolation. In this work, we showed how the latter artificially inflates performance scores, and thus can lead to reaching inaccurate conclusions about classifier quality.

Stepwise interpolation in AUARC was critical to understanding the limitations of the popular vanilla verbalization method. We proposed the \textit{verbalization logprobs} method, a simple extension of vanilla verbalization that alleviates sparsity issues, improves confidence estimates, and matches sampling-based methods at much lower cost.

\section*{Limitations}

Our verbalization logprobs approach assumes that each digit is consistently tokenized as a single token, and while this holds for the models we tested, the approach may require adaptation for tokenizers that utilize different schemes. For sampling-based approaches, we fix the number of samples at four based on prior work showing early saturation, but the optimal number may vary across models and tasks. While we focus on AUARC to demonstrate the impact of interpolation methods, similar considerations apply to other threshold-based metrics such as AUPRC, which we leave for future investigation. Finally, our proposed verbalization logprobs requires access to token probabilities; we do not provide a solution for reducing sparsity in purely verbalization-based settings where logprobs are unavailable.

%ACL currently requires all submissions to have a section titled “Limitations”, which discusses the limitations of the work. It may not contain any additional experiments, figures or analysis. It should be placed after the conclusion section and before references, without page breaks. It does not count towards the page limit.

\section*{Acknowledgements}
Elena Merdjanovska was partially funded by the Deutsche Forschungsgemeinschaft (DFG, German Research Foundation) under Germany’s Excellence Strategy – EXC 2002/1 “Science of Intelligence” – project number 390523135.


\bibliography{custom,anthology_extracted}

\clearpage
\appendix

\section{Datasets}
\label{appendix:datasets}

\begin{table*}
\small
\setlength{\tabcolsep}{4.5pt}
\centering
\begin{tabular}{lccccc}
\toprule
Dataset & Task & Text & \# Classes & Classes & Test Size \\
\midrule
SST-2 \citep{socher-etal-2013-recursive} & sentiment & movie review & 2 & [positive, negative] & 872 \\
SST-5 \citep{socher-etal-2013-recursive}  & sentiment & movie review & 5   & [very positive, positive, neutral...] & 2210\\
Yahoo \citep{zhang2015character} & topic & question title& 10  & [Society \& Culture, Health...]    & 6000\\
ESCI \citep{reddy2022shopping}  & relevance & query-product pair & 4   & [exact, substitute, complement...]   & 8604 \\
\bottomrule
\end{tabular}
\caption{Overview of the four classification datasets used in our experiments. SST-2 and SST-5 are sentiment analysis tasks of varying granularity on movie reviews. Yahoo! answers provides multi-class topic classification. Amazon ESCI tests product-query relevance classification.}
\label{tab:datasets_table}
\end{table*}

Table~\ref{tab:datasets_table} provides an overview of the four classification datasets used in our experiments. SST-2 and SST-5 represent sentiment analysis tasks of varying granularity on movie reviews. Yahoo! answers provides a multi-class topic classification task, while Amazon ESCI tests product-query relevance classification.

For computational efficiency, we sample subsets from the larger datasets. From SST-2 and SST-5 we use their full standard test splits (872 and 2,210 samples respectively). For Yahoo! answers, we randomly sample 6,000 examples from the original 60,000 test samples (10\%). For Amazon ESCI, we sample 8,604 examples from the full dataset. These sample sizes provide sufficient statistical power for our comparisons while keeping inference costs manageable across multiple methods and models.

\section{Claude 3.7 Sonnet Results}
\label{appendix:claude-results}

We present results for Claude 3.7 Sonnet to demonstrate that our findings generalize to other models. Table~\ref{tab:auarc-comparison-claude} shows the impact of interpolation method on AUARC, mirroring our main findings: consistency sampling drops from rank~1 to rank~4 when switching from linear to stepwise interpolation. Table~\ref{tab:main_results_claude} provides comprehensive metrics. Note that token logprobs are unavailable via the Bedrock API.

\begin{table}[H]
\centering
\small
\begin{tabular}{@{}lc c c c@{}}
\toprule
 & \multicolumn{3}{c}{AUARC} & \\
 \cmidrule(lr){2-4}
& linear & & step & \%t5 \\ \midrule
\noalign{\vskip -1.5mm}
 & & {\tiny ranks} & & \\
Consistency Sampling & 0.814 & {\tiny 1→4} & 0.703 & 100 \\
Vanilla Verbalization & 0.801 & {\tiny 2→3} & 0.771 & 96 \\
Verbalization Sampling & 0.795 & {\tiny 3→1} & 0.790 & 42 \\
Top-2 Verbalization & 0.795 & {\tiny 4→2} & 0.775 & 95 \\
Token Logprobs & / & & / & / \\ \bottomrule
\end{tabular}
\caption{Comparison of different interpolations when calculating AUARC for Claude 3.7 Sonnet.}
\label{tab:auarc-comparison-claude}
\end{table}

\begin{table}[H]
\centering
\small
\setlength{\tabcolsep}{5pt}
\begin{tabular}{@{}lccccc@{}}
\toprule
& AUROC & ECE & AUARC & \%t5 & Cost \\ \midrule
Random & 0.494 & 0.225 & 0.665 & 0 & 1.0$\times$ \\ \midrule
Consist. Samp. & 0.579 & 0.266 & 0.703 & 100 & 4.0$\times$ \\
Vanilla Verb. & 0.691 & 0.184 & 0.771 & 96 & 1.9$\times$ \\
Verb. Sampling & 0.705 & 0.187 & 0.790 & 42 & 7.4$\times$ \\
Top-2 Verb. & 0.710 & 0.131 & 0.775 & 95 & 2.6$\times$ \\ 
%\midrule
%\textit{Proposed} & & & & & \\
%Top-2 Margin & 0.713 & 0.136 & 0.778 & 88.5 & 2.6x \\
%Top-N Verb. & 0.713 & 0.097 & 0.776 & 89.8 & 4.1x \\
%Top-N Margin & 0.718 & 0.162 & 0.781 & 74.0 & 4.1x \\
\bottomrule
\end{tabular}
\caption{Comparison of confidence estimation methods with Claude 3.7 Sonnet (averages over our four datasets).} %\%t5 indicates concentration in top-5 most frequent confidence values. ECE is the expected calibration error. Cost is a multiple compared with predicting simply the class label (no confidence).} %Token logprobs unavailable via Bedrock API.}
\label{tab:main_results_claude}
\end{table}

\begin{table}[H]
\centering
\small
\begin{tabular}{@{}lcc@{}}
\toprule
Approach & w/o reasoning &w/ reasoning \\ \midrule
Consistency Sampling & 4.0$\times$ & 4.0$\times$ \\
Vanilla Verbalization & 1.8$\times$ & 1.3$\times$ \\
Verbalization Sampling & 7.1$\times$ & 5.0$\times$ \\
Top-2 Verbalization & 2.5$\times$ & 1.5$\times$ \\
Logprobs & 1.0$\times$ & 1.0$\times$ \\ \bottomrule
\end{tabular}
\caption{Relative cost multipliers for each confidence estimation method compared to a baseline that predicts only the class label. This uses Qwen3-32B with and without reasoning. Sampling-based approaches incur higher costs due to multiple inference calls; verbalization adds modest overhead from confidence tokens.}
\label{tab:cost-comparison-aggregated-reasoning}
\end{table}

\section{Cost Calculation}
\label{appendix:cost}

We compute relative cost as the ratio of total tokens (input + output) compared to a baseline that predicts only the class label without confidence estimation. Table~\ref{tab:cost-comparison-aggregated-reasoning} shows the cost multipliers for each method using Qwen3-32B with and without reasoning. Sampling-based approaches incur higher costs due to multiple inference calls. Verbalization adds modest overhead from the confidence tokens. Token logprobs has no additional cost as it uses the same inference call.

\section{Reasoning Results}
\label{appendix:reasoning}

Table~\ref{tab:reasoning-aggregated-results} presents results with reasoning mode enabled (1024 reasoning budget tokens for Claude). Sparsity patterns persist even with reasoning, confirming that the phenomenon is not limited to direct prediction settings.

\begin{table*}
\centering
\small
\setlength{\tabcolsep}{3.5pt}
\begin{tabular}{@{}clllllllllll@{}}
\toprule
\multicolumn{1}{l}{} &  & \multicolumn{5}{c}{\textit{Qwen3-32B}} & \multicolumn{5}{c}{\textit{Claude 3.7 Sonnet}} \\
\multicolumn{1}{l}{} & Approach & \multicolumn{1}{c}{AUROC↑} & \multicolumn{1}{c}{ECE↓} & \multicolumn{1}{c}{AUARC↑} & \multicolumn{1}{c}{\%t5↓} & \multicolumn{1}{c}{Cost↓} & \multicolumn{1}{c}{AUROC↑} & \multicolumn{1}{c}{ECE↓} & \multicolumn{1}{c}{AUARC↑} & \multicolumn{1}{c}{\%t5} & \multicolumn{1}{c}{Cost↓} \\
\cmidrule(lr){1-2} \cmidrule(lr){3-7}  \cmidrule(lr){8-12}
\multirow{6}{*}{\rotatebox[origin=c]{90}{baselines}} & Random confidences & 0.498 & 0.223 & 0.674 & 0 & 1.00$\times$ & 0.507 & 0.225 & 0.716 & 0 & 1.00$\times$ \\
 & Consistency sampling & 0.622 & 0.223 & 0.706 & 100 & 4.08$\times$ & 0.569 & 0.238 & 0.732 & 100 & 4.00$\times$ \\
 & Vanilla verbalization & 0.659 & 0.217 & 0.739 & 89 & 1.21$\times$ & 0.707 & 0.182 & 0.794 & 90 & 1.30$\times$ \\
 & Verbalization sampling & \textbf{0.688} & 0.221 & \textbf{0.768} & 43 & 4.87$\times$ & 0.719 & 0.183 & \textbf{0.807} & 49 & 5.19$\times$ \\
 & Top-2 verbalization & 0.682 & \textbf{0.174} & 0.757 & 85 & 1.60$\times$ & \textbf{0.732} & \textbf{0.136} & 0.803 & 85 & 1.50$\times$ \\
 & Token Logprobs & 0.529 & 0.280 & 0.711 & 0 & 1.00$\times$ & / & / & / & / & / \\ \bottomrule
\end{tabular}
\caption{Confidence estimation results with reasoning mode enabled. Claude uses 1024 reasoning budget tokens. Sparsity (\%t5) remains high for verbalization approaches even with reasoning, confirming that the phenomenon persists across inference settings. Token logprobs are unavailable for Claude via Bedrock API.}
\label{tab:reasoning-aggregated-results}
\end{table*}

\section{Sparsity Investigation for Alternative Confidence Range~0--9}
\label{appendix:alternative-confidence-range}

In our main experiments, we instruct the model to verbalize its confidence from 0--100, which is arguably the most natural range as it aligns well with percentages and is consistent with prior work. In this appendix, we present results using an alternative confidence range of 0--9 to verify that our findings are not an artifact of the chosen confidence range.

\begin{figure*}
    \centering
    \small
    \begin{subfigure}{0.47\textwidth}
        \centering
        \includegraphics[width=\linewidth]{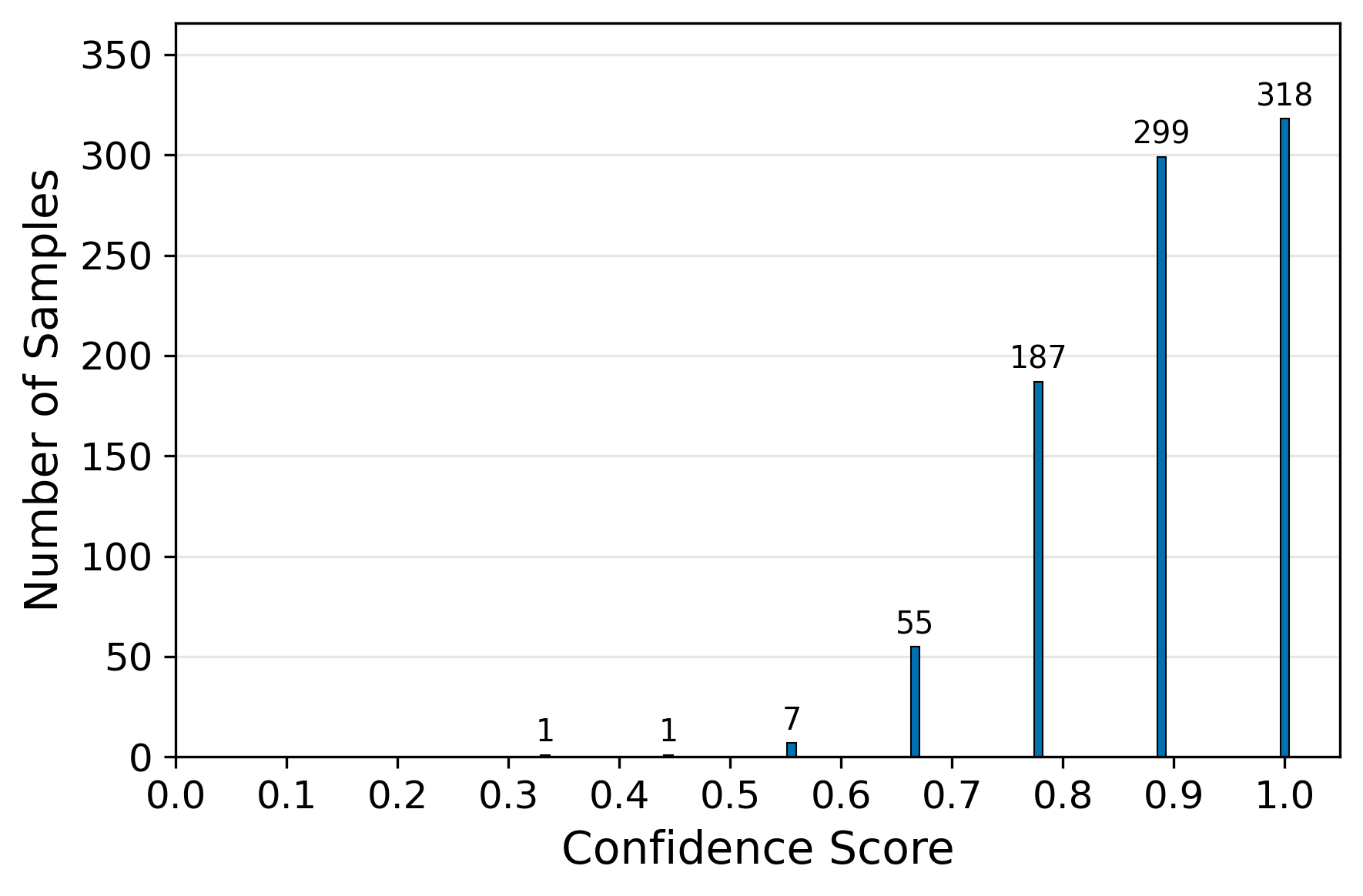}
        \caption{Histogram showing all confidence scores on SST-2 with an alternative confidence range of 0--9}
        \label{fig:practical_sparsity_alternative_range_histogram}
    \end{subfigure}
    \hfill
    \begin{subfigure}{0.47\textwidth}
        \centering
        \includegraphics[width=\linewidth]{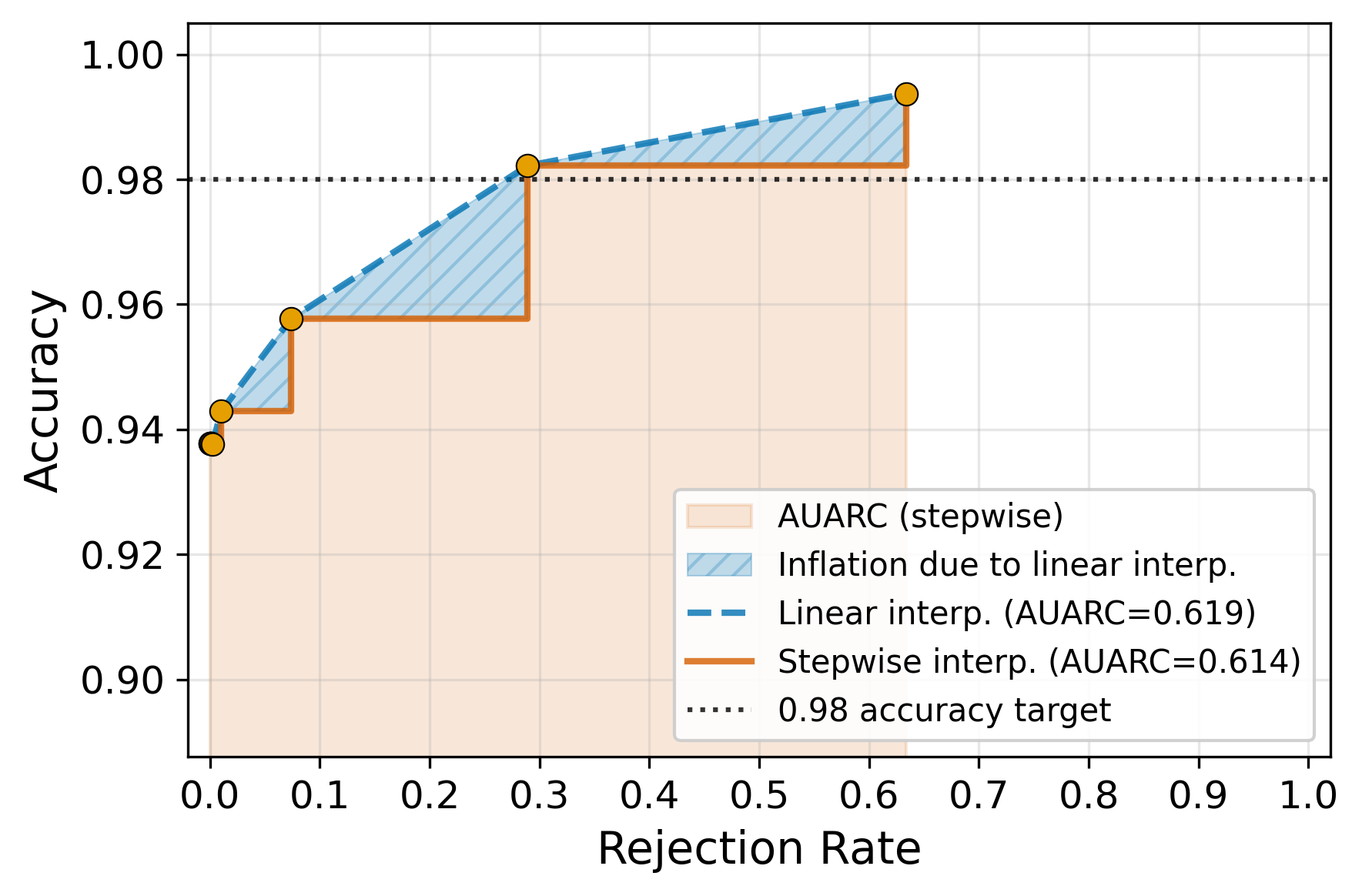}
        \caption{Accuracy-rejection curve on SST-2 with an alternative confidence range of 0--9}
        \label{fig:practical_sparsity_alternative_range_arc}
    \end{subfigure}
    \caption{\textbf{Sparse confidences for confidence scoring range 0--9.} 
    (a)~Histogram of vanilla verbalized confidences on SST-2 with Qwen3-32B using the confidence range 0--9 (normalized to 0--1). Only seven unique values appear, with the majority concentrated at $\frac{8}{9}$ and $\frac{9}{9}$.
    (b)~Accuracy-rejection curve demonstrating limited threshold choices due to sparsity.}    
    \label{fig:practical_sparsity_alternative_range}
\end{figure*}

We run vanilla verbalization and verbalization sampling with this new confidence range, following the setup in Section~\ref{sec:sparsity-limitation}. For direct comparison with Figure~\ref{fig:practical_sparsity}, we show in Figure~\ref{fig:practical_sparsity_alternative_range} confidence histograms and accuracy-rejection curves for the same dataset and model. As expected, we again find extreme sparsity of confidence values---the narrower range offers even fewer distinct values. Consistent with our prior findings, the majority of confidence scores are concentrated in just two values, $\frac{8}{9}$ and $\frac{9}{9}$.

Table~\ref{tab:auarc-alternative-range} compares both confidence ranges across both models. As expected, concentration in the top-5 values (\%t5) increases substantially with the narrower range, while AUARC scores remain comparable. Importantly, linear interpolation continues to inflate AUARC scores for both ranges. Our results confirm that both the sparsity limitation and the inflation from linear interpolation persist with the alternative range.

\begin{table}[]
\centering
\small
\begin{tabular}{@{}lc c c@{}}
\toprule
 & \multicolumn{2}{c}{AUARC} & \\
 \cmidrule(lr){2-3}
 & linear & step & \%t5 \\ \midrule
\textit{Qwen3 32B} \\ \cmidrule{1-4}
Vanilla verbalization & & & \\
~~~\emph{conf.\ range 0--100} & 0.758 & 0.713 & 92 \\
~~~\emph{conf.\ range 0--9} & 0.761 & 0.716 & 99 \\
Verbalization sampling & & & \\
~~~\emph{conf.\ range 0--100} & 0.741 & 0.734 & 45 \\
~~~\emph{conf.\ range 0--9} & 0.754 & 0.728 & 66 \\ \midrule
\textit{Claude 3.7 Sonnet} \\ \cmidrule{1-4}
Vanilla verbalization & & & \\
~~~\emph{conf.\ range 0--100} & 0.801 & 0.771 & 96 \\
~~~\emph{conf.\ range 0--9} & 0.805 & 0.773 & 99 \\
Verbalization sampling & & & \\
~~~\emph{conf.\ range 0--100} & 0.795 & 0.790 & 42 \\
~~~\emph{conf.\ range 0--9} & 0.800 & 0.788 & 59 \\
\bottomrule
\end{tabular}
\caption{Impact of confidence scoring range on AUARC and sparsity. Scores are averages over SST-2, SST-5, Amazon ESCI, and Yahoo! topic classification (see Table~\ref{tab:datasets_table} for dataset details).
\%t5 refers to the concentration in top-5 confidence values. The narrower range 0--9 increases concentration substantially while AUARC scores remain comparable.}
\label{tab:auarc-alternative-range}
\end{table}

\section{Confidence Estimation Details}
\label{appendix:example-prompts}

Below we describe the confidence estimation methods included in our experimental comparison. Table \ref{tab:example_prompts} shows example prompts for each approach on the SST-2 dataset. Claude uses the Bedrock API, and Qwen3-32B uses vLLM. We generate tokens at temperature=0 if not otherwise mentioned with 1000 maximum tokens.  

\begin{itemize}
    \item \textbf{Vanilla verbalization} \citep{tian-etal-2023-just, xiong2024can}: The model is prompted to output a self-reported confidence score as a percentage between 0 and 100 (e.g., ``Label and Confidence: positive, 85\%'').
    \item \textbf{Top-K verbalization} \citep{tian-etal-2023-just}: The model outputs its $k$ best guesses along with confidence scores for each. We use $k=2$ and take the confidence of the top-ranked guess.
    \item \textbf{Consistency sampling} \citep{xiong2024can}: The same prompt is executed multiple times without requesting confidence scores. Confidence is computed as the frequency of the most common predicted label. We sample 4 responses~\citep{xiong2024can} at temperature 1.0~\citep{zhang-etal-2024-calibrating}.
    \item \textbf{Verbalization sampling} \citep{xiong2024can}: Similar to consistency sampling, but using the vanilla verbalization prompt. Confidence is the mean of the sampled verbalized values. We use the same sampling parameters (4 responses, temperature 1.0).
    \item \textbf{Token logprobs} \citep{huang2025look}: We use a prompt without explicit confidence requests and compute confidence from the log-probability of the predicted class label token. For multi-token labels, we average the log-probabilities. This approach requires API access to token probabilities.
    \item \textbf{Random baseline}: We assign uniformly random confidence values between 0 and 1 to establish a lower bound for comparison and to calculate relative cost increases.
\end{itemize}

\begin{table*}
\small
\centering
\begin{tabular}{l p{0.8\textwidth}}
\toprule
\textbf{Prompt Type} & Baseline prompt without confidence \\ \midrule
\textbf{Example Prompt} & \begin{verbatim} Given the sentence, assign a sentiment label from ['negative', 'positive'].

Use the following format:

```Label [ONLY the sentiment label; not a complete sentence]```

Only the label, don't give me the explanation.

Sentence: too much of the humor falls flat .
\end{verbatim} 
\\

\textbf{Used in} & Random, Logprobs, Consistency Sampling \\

\midrule
\midrule

\textbf{Prompt Type} & Vanilla Verbalization \\ \midrule
\textbf{Example Prompt} & \begin{verbatim}
Given the sentence, assign a sentiment label from ['negative', 'positive'] and your 
confidence in this answer. The confidence indicates how likely you 
think your answer is true.

Use the following format:
```
Label and Confidence (0-100): [ONLY the sentiment label; not a complete sentence], 
[Your confidence level, please only include the number in the range of 0-100]%
```

Only the label and confidence, don't give me the explanation.

Sentence: too much of the humor falls flat .
\end{verbatim}
\\
\textbf{Used in} & Vanilla Verbalization, Verbalization Sampling\\
\midrule
\midrule
\textbf{Prompt Type} & Top-2 Verbalization\\
\midrule

\textbf{Example Prompt} &\begin{verbatim}
Given the sentence, assign a sentiment label from ['negative', 'positive'].
Provide your 2 best guesses and the probability that each is correct (0\% to 100\%). 
Give only the sentiment label and probabilities, no other words or explanation.

Example:

G1: <ONLY the sentiment label of first most likely guess; not a complete sentence, 
the guess!>
P1: <ONLY the probability that G1 is correct, without any extra commentary whatsoever; 
the probability!>
...
G2: <ONLY the sentiment label of 2-th most likely guess>
P2: <ONLY the probability that G2 is correct, without any extra commentary whatsoever; 
the probability!>

Sentence: too much of the humor falls flat .
\end{verbatim}
\\
\textbf{Used in} & Top-2 Verbalization \\

\bottomrule
\end{tabular}
\caption{Example Prompts for SST-2, with a list of approaches that uses each prompt. \label{tab:example_prompts}}
\end{table*}

\section{Linear Interpolation in Prior Work}
\label{sec:prior-work-implementation}

We verified that prior work uses linear interpolation when calculating AUARC, by inspecting publicly available code. We found that researchers commonly use \texttt{scikit-learn}'s~\citep{scikit-learn} \texttt{auc()} function~\citep{nguyen-etal-2025-beyond,lin-etal-2024-contextualized,lin2024generating}, which uses trapezoidal integration (i.e. linear interpolation) per its public documentation. It is not surprising that the use of this function became the default; unlike AUROC and AUPRC---which have dedicated \texttt{scikit-learn} functions with correct interpolation: \texttt{roc\_auc\_score()} and \texttt{average\_precision\_score()}, respectively---there is no specialized function for AUARC.%, causing researchers to default to \texttt{auc()}. 

%It is important to raise the awareness of this issue, and it would be useful to incorporate stepwise interpolation for AUARC computation in, e.g., the uncertainty evaluation tool LM-Polygraph~\citep{vashurin-etal-2025-benchmarking}. 
Other implementations compute AUARC without calling \texttt{auc()} but still approximate linear interpolation. For instance, LM-Polygraph~\citep{vashurin-etal-2025-benchmarking} can compute AUARC (via the Prediction-Rejection Ratio---PRR, equivalent to AUARC for classification tasks) but does so by thresholding at every prediction in order of confidence scores, regardless of whether multiple predictions share the same confidence. With random tie breaking, it approximates linear interpolation between the confidence-thresholded anchors. 

While these implementations are valid for continuously-distributed confidence scores, they compute overly optimistic AUARC scores for sparse verbalized confidences.

\end{document}